\documentclass{article} %
\usepackage{iclr2027_conference,times}

\usepackage{amsmath,amsfonts,bm}

\def\eqref#1{equation~\ref{#1}}

\def\1{\bm{1}}

\DeclareMathAlphabet{\mathsfit}{\encodingdefault}{\sfdefault}{m}{sl}
\SetMathAlphabet{\mathsfit}{bold}{\encodingdefault}{\sfdefault}{bx}{n}

\def\sF{{\mathbb{F}}}

\def\sR{{\mathbb{R}}}

\newcommand{\E}{\mathbb{E}}

\DeclareMathOperator*{\argmax}{arg\,max}

\usepackage{hyperref}
\usepackage{cleveref}
\usepackage{url}

\usepackage{amsmath}
\usepackage{bm}
\usepackage{xcolor}
\usepackage{float}
\usepackage{algorithm}
\usepackage{algorithmic}
\usepackage{natbib}
\usepackage{graphicx}
\usepackage{booktabs}

\usepackage[T1]{fontenc}
\usepackage{tikz}
\usepackage{subcaption}
\usepackage{wrapfig}
\usepackage{xcolor}
\usepackage{booktabs}
\usepackage{multirow}

\title{Fork-dLLM: Avoiding the Flexibility Trap in Diffusion Language Models}

\author{Stipe Frkovi\'{c}$^{1,\dagger}$, Metod Jazbec$^{2,*,\dagger}$, Christian A. Naesseth$^{2,*}$ \\
$^{1}$University of Amsterdam \quad $^{2}$UvA-Bosch Delta Lab, University of Amsterdam
}

\def\preprint{} %
\ifdefined\preprint\iclrfinalcopy\fi
\begin{document}

\maketitle
\ifdefined\preprint\lhead{Preprint}{\renewcommand{\thefootnote}{*}\footnotetext{Equal supervision.\enspace$^{\dagger}$Correspondence to: {\fontsize{7.5}{9}\selectfont\texttt{stipe.frkovic@student.uva.nl}}, {\fontsize{7.5}{9}\selectfont\texttt{m.jazbec@uva.nl}}.}}\fi

\begin{abstract}
Masked diffusion language models (dLLMs) have shown strong potential for faster inference through parallel token generation when combined with confidence-based samplers. However, recent work has shown that such methods can defer unmasking high-entropy \emph{fork} positions at which multiple plausible continuations exist. This results in reduced generation diversity, as shown by worse pass@$k$ scaling, and limits gains obtainable from RL post-training. To avoid this \emph{flexibility trap}, prior work advocated for autoregressive (AR) sampling. Here, we show that discarding confidence-based sampling is unnecessary and, once inference cost is taken into account, wasteful. We first propose \emph{Fork-dLLM}, a simple hybrid sampler that uses AR-style ordering only at uncertain fallback steps while retaining parallel generation otherwise. We then extend the same principle to post-training with \emph{ForkGRPO}, which uses Fork-dLLM rollouts and applies the GRPO objective only at fallback steps, preserving exact policy-likelihood ratios while substantially reducing rollout and optimization cost. In our experiments, Fork-dLLM matches the strong pass@$k$ scaling of AR sampling while being $2$--$3\times$ more efficient, and ForkGRPO achieves downstream performance comparable to or better than AR-based GRPO baselines at a substantially lower training cost. %
\end{abstract}

\section{Introduction}

\definecolor{geBase}{HTML}{4D4D4D}   %
\definecolor{geFork}{HTML}{6A3D9A}   %
\definecolor{geJust}{HTML}{FF4500}   %
\definecolor{geJustF}{HTML}{D95F02}  %

\def\xsc{0.010547}   %
\def\ysc{35}         %
\def\yzero{0.80}     %
\def\xo{0}           %

\providecommand{\gept}[2]{({#1*\xsc},{(#2-\yzero)*\ysc})}

\ifdefined\preprint\def\figonewraplines{23}\else\def\figonewraplines{25}\fi
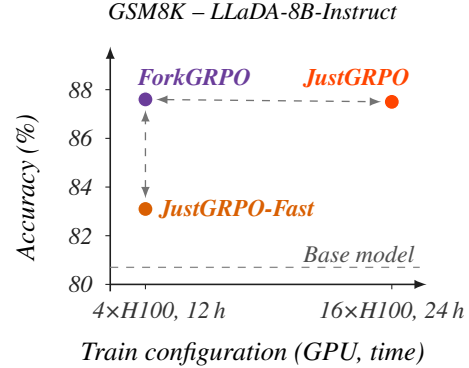
\begin{wrapfigure}[\figonewraplines]{r}{0.44\textwidth}
\ifdefined\preprint\vspace{-40pt}\else\vspace{-21pt}\fi
\noindent
\begin{tikzpicture}[scale=0.92,
    dot/.style={
        circle,
        draw=white,
        line width=0.5pt,
        minimum size=2.0mm,
        inner sep=0pt
    },
    tick/.style={
        font=\footnotesize\itshape,
        text=black!88
    },
    axlab/.style={
        font=\normalsize
    },
]

  \draw[black!88, line width=0.6pt, -latex]
    (\xo,-0.10) -- (\xo,3.42);

  \draw[black!88, line width=0.6pt, -latex]
    (\xo,0) -- (4.55,0);

  \foreach \a/\l in {0.80/80, 0.82/82, 0.84/84, 0.86/86, 0.88/88} {
    \draw[black!88]
      (\xo,{(\a-\yzero)*\ysc})
      -- ({\xo-0.07},{(\a-\yzero)*\ysc});

    \node[tick, anchor=east]
      at ({\xo-0.12},{(\a-\yzero)*\ysc})
      {\l};
  }

  \foreach \h in {48,384} {
    \draw[black!88]
      ({\h*\xsc},0)
      -- ({\h*\xsc},-0.07);
  }

  \node[
      font=\footnotesize\itshape,
      text=black!80,
      anchor=north
    ]
    at ({48*\xsc+0.20},-0.12)
    {4\texttimes H100, 12\,h};

  \node[
      font=\footnotesize\itshape,
      text=black!80,
      anchor=north
    ]
    at ({384*\xsc},-0.12)
    {16\texttimes H100, 24\,h};

  \node[axlab, anchor=north]
    at (2.05,-0.70)
    {\textit{Train configuration (GPU, time)}};

  \node[
      font=\normalsize,
      rotate=90,
      anchor=south
    ]
    at ({\xo-0.88},1.40)
    {\textit{Accuracy (\%)}};

  \node[
      font=\small,
      anchor=south
    ]
    at (2.05,3.68)
    {\textit{GSM8K} -- \textit{LLaDA-8B-Instruct}};

  \draw[
      geBase!65,
      dashed,
      line width=0.5pt
    ]
    (\xo,{(0.807-\yzero)*\ysc})
    -- (4.45,{(0.807-\yzero)*\ysc});

  \node[
      font=\footnotesize\itshape,
      text=geBase,
      anchor=south east,
      inner sep=1pt
    ]
    at (4.42,{(0.807-\yzero)*\ysc+0.03})
    {Base model};

  \draw[
      black!55,
      dashed,
      line width=0.5pt,
      latex-latex,
      shorten <=4pt,
      shorten >=4pt
    ]
    \gept{48}{0.876}
    --
    \gept{384}{0.875};

  \draw[
      black!55,
      dashed,
      line width=0.5pt,
      latex-latex,
      shorten <=4pt,
      shorten >=4pt
    ]
    \gept{48}{0.831}
    --
    \gept{48}{0.876};

  \node[dot, fill=geFork]
    at \gept{48}{0.876} {};

  \node[dot, fill=geJust]
    at \gept{384}{0.875} {};

  \node[dot, fill=geJustF]
    at \gept{48}{0.831} {};

  \node[
      anchor=south west,
      inner sep=1pt
    ]
    at ({48*\xsc-0.15},{(0.876-\yzero)*\ysc+0.16})
    {\footnotesize\bfseries\itshape
     \textcolor{geFork}{ForkGRPO}};

  \node[
      anchor=west,
      inner sep=1pt
    ]
    at ({48*\xsc+0.16},{(0.831-\yzero)*\ysc})
    {\footnotesize\bfseries\itshape
     \textcolor{geJustF}{JustGRPO-Fast}};

  \node[
      anchor=south east,
      inner sep=1pt
    ]
    at ({384*\xsc+0.30},{(0.875-\yzero)*\ysc+0.16})
    {\footnotesize\bfseries\itshape
     \textcolor{geJust}{JustGRPO}};

\end{tikzpicture}

\caption{
GSM8K accuracy across post-training methods, evaluated with Fast-dLLM at $\lambda=0.95$ and temperature $\tau=0$. The dashed gray line denotes the performance of the base LLaDA-8B-Instruct model. ForkGRPO and JustGRPO-Fast are trained by us; for JustGRPO, we evaluate the released checkpoint from \citet{ni2026flexibility}. Under the same training budget, ForkGRPO substantially outperforms JustGRPO-Fast and reaches accuracy comparable to JustGRPO despite using $8\times$ fewer H100-hours for post-training.
}
\label{fig:grpo-efficiency}

\vspace{-\intextsep}
\end{wrapfigure}

Masked diffusion language models (dLLMs) have emerged as a promising alternative to autoregressive (AR) language models, with the potential for faster inference through parallel token generation \citep{nie2026large, ye2025dream,arriola2025block,bie2025llada2,cheng2026sdar}. To realize this potential, a crucial component is confidence-based sampling \citep{wu2025fastdllmtrainingfreeaccelerationdiffusion,ben-hamuAcceleratedSamplingMasked2025,kim2025train,jazbec2025learning}, accelerating generation by committing multiple high-confidence tokens in parallel.

However, recent work has shown that confidence-based sampling can substantially reduce diversity across generations \citep{ni2026flexibility,olausson2026tale}, a phenomenon termed the \emph{flexibility trap}. By postponing uncertain, high-entropy \emph{fork} positions \citep{wang2026beyond}, confidence-based samplers allow surrounding context to be resolved first, reducing uncertainty at meaningful branching points before they are sampled. This leads to worse pass@$k$ scaling than left-to-right AR sampling and limits the gains of RL post-training. \citet{ni2026flexibility} therefore advocate for AR sampling, which generates only one token at a time and thus sacrifices the parallelism and corresponding efficiency gains of dLLMs, making it an expensive solution to the diversity problem.

We therefore ask: \emph{Is it necessary to abandon confidence-based sampling entirely to avoid the flexibility trap?} We show that it is not. Our key observation is that popular confidence-based samplers such as Fast-dLLM \citep{wu2025fastdllmtrainingfreeaccelerationdiffusion} already contain a mechanism for identifying sampling steps in which fork positions are likely to occur: when no position exceeds the confidence threshold $\lambda$, all remaining predictions are relatively uncertain. We use these \emph{fallback} steps as a simple proxy for fork-containing steps and introduce \emph{Fork-dLLM} (\Cref{alg:fork-dllm}), which switches to AR-style sampling only at these steps by unmasking the leftmost masked position. In all other steps, Fork-dLLM is identical to Fast-dLLM and commits multiple tokens in parallel.

We further show that the same idea naturally extends to RL post-training for dLLMs \citep{zhao2025d}. Existing approaches that avoid likelihood approximations (see \Cref{sec:rel-work}), such as JustGRPO \citep{ni2026flexibility}, rely on AR rollouts and therefore sacrifice the generation efficiency of dLLMs. We instead introduce \emph{ForkGRPO}, which generates rollouts with Fork-dLLM and applies the GRPO objective only at fallback steps. Since a single token is committed at these steps, ForkGRPO preserves exact policy-likelihood ratios while retaining parallel generation throughout the rest of the trajectory. In summary, our main contributions are:
\begin{itemize}
\item We introduce \emph{Fork-dLLM}, a simple modification to the Fast-dLLM sampler \citep{wu2025fastdllmtrainingfreeaccelerationdiffusion} that applies AR-style ordering only at uncertain fallback steps, requiring no training, additional model evaluations, or new hyperparameters (\Cref{sec:methods-forkdllm}). Across mathematics and code benchmarks, Fork-dLLM matches the pass@$k$ scaling of AR sampling while requiring $2$--$3\times$ fewer NFEs (\Cref{sec:sampling}).

\item We show that, when tuned, other confidence-based samplers \citep{wu2025fastdllmtrainingfreeaccelerationdiffusion, olausson2026tale} also retain strong pass@$k$ scaling at a fraction of the AR cost, demonstrating that escaping the flexibility trap does not require abandoning confidence-based parallel decoding (\Cref{sec:sampling}).

\item We introduce \emph{ForkGRPO}, which extends Fork-dLLM to RL post-training by applying GRPO only at the fallback steps of its rollouts (\Cref{sec:methods-forkgrpo}). ForkGRPO preserves exact policy-likelihood ratios while substantially reducing NFEs during both rollout generation and policy optimization. It clearly outperforms AR-based GRPO under the same compute budget, and matches it at a fraction of the training cost (\Cref{fig:grpo-efficiency} \& \Cref{sec:grpo}).

\end{itemize}

\section{Background}

\paragraph{Training masked dLLMs.}
Let $[V] := \{1, ..., V\}$ represent a vocabulary of size $V$ and $\bm{x}_0=(x_0^1,\ldots,x_0^L)\sim p_{\mathrm{data}}$ denote a clean sequence of length $L$, where $x_0^{\ell} \in [V]$. Masked dLLMs \citep{sahoo2024simple, shi2024simplified, ou2025} define a forward process that progressively corrupts $\bm{x}_0$ by replacing tokens with an absorbing \texttt{[MASK]} token. Concretely, a (continuous) timestep $s\sim\mathcal{U}[0,1]$ is sampled and used to independently mask every position with probability $s$,
\begin{equation*}
p_s(\bm{x}_s\mid\bm{x}_0)
=
\prod_{k=1}^{L}
\left[
s \cdot\,\mathbf{1}\!\left(x_s^k=\texttt{[MASK]}\right)
+
(1-s) \cdot\,\mathbf{1}\!\left(x_s^k=x_0^k\right)
\right].
\end{equation*}
A bidirectional transformer $q_\theta$ is then trained to reconstruct the clean tokens at masked positions using the cross-entropy:
\begin{equation*}
\mathcal{L}(\theta)
=
-\mathbb{E}_{s,\bm{x}_0,\bm{x}_s}
\left[
\frac{1}{s}
\sum_{k=1}^{L}
\mathbf{1}\!\left(x_s^k=\texttt{[MASK]}\right) \cdot
\log q_\theta^k(x_0^k\mid\bm{x}_s)
\right].
\label{eq:dllm_training}
\end{equation*}
Thus, $q_\theta^k(\cdot\mid\bm{x}_s)$ predicts the distribution over the clean token at position $k$ conditioned on the entire partially masked sequence. Such models have recently been scaled to billions of parameters \citep{nie2026large, nie2026improved, ye2025dream, bie2025llada2,bethune2026design}.

\paragraph{Sampling in masked dLLMs.}  At inference time, generation proceeds through iterative unmasking over discrete sampling steps indexed by $t$, until all positions have been unmasked \citep{zheng2025masked,nie2026large}. Let $\bm{x}_t=(x_t^1,\ldots,x_t^L)$ denote the current partially masked sequence and $\mathcal{M}_t=\{k\in[L]\mid x_t^k=\texttt{[MASK]}\}$ the set of still-masked positions. At each sampling step, the model predicts a distribution $q_\theta^k(\cdot\mid\bm{x}_t)$ for every $k\in\mathcal{M}_t$, from which we sample a candidate token $ x_0^k \sim q_\theta^k(\cdot\mid\bm{x}_t;\tau)$, with $\tau \ge 0$ denoting the sampling temperature. A sampling strategy then selects a subset of proposed tokens $\mathcal{U}_t\subseteq\mathcal{M}_t$ to commit/unmask:
\begin{equation*} x_{t-1}^k = \begin{cases} x_0^k, & k\in\mathcal{U}_t,\\ x_t^k, & \text{otherwise}. \end{cases} \end{equation*} 
The choice of $\mathcal{U}_t$ determines the generation order and, crucially, how many tokens can be generated in parallel. Confidence-based samplers exploit this flexibility to accelerate generation. One popular instantiation is Fast-dLLM \citep{wu2025fastdllmtrainingfreeaccelerationdiffusion} in which each sampled token is assigned the confidence $c_t^k=q_\theta^k(x_0^k\mid\bm{x}_t)$ and then all positions with confidence above a threshold $\lambda$ are unmasked:
\begin{equation*} 
\mathcal{U}_t=\{k\in\mathcal{M}_t:c_t^k>\lambda\}. \end{equation*} 
If no position exceeds the threshold (i.e., $\mathcal{U}_t = \emptyset$), Fast-dLLM instead commits the highest-confidence position $\mathcal{U}_t=\argmax_k c_t^k$, ensuring that at least one token is generated at every step (\Cref{alg:fast-dllm}).

\begin{figure}[t]
\vspace{-15pt}
\begin{minipage}[t]{0.48\textwidth}
\begin{algorithm}[H] \caption{Fast-dLLM \citep{wu2025fastdllmtrainingfreeaccelerationdiffusion}} \label{alg:fast-dllm} \begin{algorithmic}[1]
\setlength{\baselineskip}{1.25\baselineskip}
\STATE \textbf{Input:} $\bm{x}_t$, $q_\theta$, $\lambda$, $\tau$ \STATE \textbf{Output:} $\bm{x}_{t-1}$ \STATE $\mathcal{M}_t \gets \{k \in [L] \mid x_t^k = \texttt{[MASK]}\}$ \STATE $x_0^k \sim q_\theta^k(\cdot \mid \bm{x}_t; \tau)$ \STATE $c_t^k \gets q_\theta^k(x_0^k \mid \bm{x}_t)$ \STATE $\mathcal{U}_t \gets \{k \in \mathcal{M}_t : c_t^k > \lambda\}$ \IF{$\mathcal{U}_t = \emptyset$} \STATE $\mathcal{U}_t \gets \{\arg\max_{k \in \mathcal{M}_t} c_t^k\}$ \ENDIF \STATE $x_{t-1}^k := \begin{cases} x_0^k, & k \in \mathcal{U}_t \\ x_t^k, & \text{else} \end{cases}$ \end{algorithmic} 
\end{algorithm}
\end{minipage}
\hfill
\begin{minipage}[t]{0.48\textwidth}
\begin{algorithm}[H] \caption{Fork-dLLM (ours)} \label{alg:fork-dllm} \begin{algorithmic}[1]
\setlength{\baselineskip}{1.25\baselineskip}
\STATE \textbf{Input:} $\bm{x}_t$, $q_\theta$, $\lambda$, $\tau$ \STATE \textbf{Output:} $\bm{x}_{t-1}$ \STATE $\mathcal{M}_t \gets \{k \in [L] \mid x_t^k = \texttt{[MASK]}\}$ \STATE $x_0^k \sim q_\theta^k(\cdot \mid \bm{x}_t; \tau)$ \STATE $c_t^k \gets q_\theta^k(x_0^k \mid \bm{x}_t)$ \STATE $\mathcal{U}_t \gets \{k \in \mathcal{M}_t : c_t^k > \lambda\}$ \IF{$\mathcal{U}_t = \emptyset$} \STATE \colorbox{cyan!15}{$
\mathcal{U}_t \gets \{\operatorname{leftmost}(\mathcal{M}_t)\}$} \ENDIF
\STATE $x_{t-1}^k := \begin{cases} x_0^k, & k \in \mathcal{U}_t \\ x_t^k, & \text{else} \end{cases}$ \end{algorithmic} 
\end{algorithm}
\end{minipage}
\end{figure}

\paragraph{GRPO for masked dLLMs.}
RL methods developed for autoregressive LLMs, such as Group Relative Policy Optimization (GRPO; \citealp{shao2024deepseekmath}), are difficult to apply to dLLMs, since their generation process lacks the left-to-right factorization of the policy likelihood. Early approaches therefore rely on likelihood approximations \citep{zhao2025d,tang2026wd1,gong2026diffucoder}.
In particular, \emph{d1} \citep{zhao2025d} proposes diffu-GRPO, which approximates the sequence likelihood using a mean-field factorization across positions and estimates all per-token likelihoods with a single dLLM forward pass on a partially masked prompt and fully masked completion.

More recently, JustGRPO \citep{ni2026flexibility} avoids this approximation by generating rollouts autoregressively and using a \emph{trajectory}-level policy likelihood \citep{chen2025dultra,turok2026duel}, achieving strong results on mathematical reasoning and code generation. AR decoding simply commits masked positions from left to right. For a generated sequence $\bm{x}_0$, let
\[
\tilde{\bm{x}}^{\,k}
=
(x_0^1,\ldots,x_0^{k-1},
\texttt{[MASK]},\ldots,\texttt{[MASK]}),
\qquad k=1,\ldots,L,
\]
denote the partially masked state immediately before position $k$ is generated. The likelihood of the resulting AR trajectory $\tilde{\bm{x}}^{\,1:L}$ then factorizes exactly as
\[
q_\theta^{\mathrm{AR}}
\bigl(\tilde{\bm{x}}^{\,1:L}\mid c\bigr)
=
\prod_{k=1}^{L}
q_\theta^k
\bigl(x_0^k\mid c,\tilde{\bm{x}}^{\,k}\bigr).
\]

For each prompt $c\sim\mathcal{D}$, JustGRPO then follows the standard GRPO recipe: it samples a group of $G$ trajectories $\{\tilde{\bm{x}}_{i}^{\,1:L}\}_{i=1}^G$ using the current policy $q_{\theta_{\mathrm{old}}}^{\mathrm{AR}}$ and optimizes the clipped surrogate objective
\begin{equation}
\mathcal{J}_{\mathrm{JustGRPO}}(\theta)
=
\underset{
\substack{
c\sim\mathcal{D},\\
\{\tilde{\bm{x}}_{i}^{\,1:L}\}_{i=1}^G
\sim q_{\theta_{\mathrm{old}}}^{\mathrm{AR}}(\cdot\mid c)
}
}{\E}\!\left[
\frac{1}{G}\sum_{i=1}^G
\frac{1}{L}\sum_{k=1}^L
\min\!\left(
\rho_{i,k}\hat A_i,\,
\operatorname{clip}(\rho_{i,k},1-\epsilon,1+\epsilon)\hat A_i
\right)
\right],
\label{eq:justgrpo}
\end{equation}
where the policy likelihood ratios are
\[
\rho_{i,k}
=
\frac{
q_\theta^k
\bigl(x_{i,0}^k\mid c,\tilde{\bm{x}}_{i}^{\,k}\bigr)
}{
q_{\theta_{\mathrm{old}}}^k
\bigl(x_{i,0}^k\mid c,\tilde{\bm{x}}_{i}^{\,k}\bigr)
},
\]
and $\hat A_i=(r_i-\mu_G)/\sigma_G$ is the group-normalized advantage, with $r_i = r(\bm{x}_{i,0})$ the reward of rollout $i$ and $\mu_G$ and $\sigma_G$ the mean and standard deviation of the rewards within the group. Thus, unlike diffu-GRPO, JustGRPO uses exact policy likelihoods, but incurs substantial training cost from both autoregressive rollouts in the GRPO outer loop and per-position likelihood evaluations in the inner loop.

\section{Methods}
\label{sec:methods}
We now introduce our approach to addressing the flexibility trap in dLLMs without resorting to AR sampling rollouts or ELBO-style likelihood approximations. We first propose \emph{Fork-dLLM}, a simple modification of the Fast-dLLM sampler that changes only its fallback rule (\Cref{sec:methods-forkdllm}) at no additional cost. We then extend the same idea to post-training with \emph{ForkGRPO}, which uses Fork-dLLM rollouts and applies the GRPO objective only at fallback steps (\Cref{sec:methods-forkgrpo}).

\subsection{Fork-dLLM}
\label{sec:methods-forkdllm}

Our key observation is that the Fast-dLLM fallback condition $\mathcal{U}_t = \emptyset$ naturally identifies steps at which confidence-based ordering is most harmful for generation diversity. When no position exceeds $\lambda$, the model has no sufficiently confident prediction; nevertheless, Fast-dLLM continues to use confidence to determine which position to commit, favoring the easiest remaining prediction and potentially further postponing uncertain, high-entropy \emph{fork} positions. This is particularly undesirable because such uncertain positions can carry a disproportionate amount of information \citep{fu2025bitsroundsparalleldecoding}.
Indeed, as we show in \Cref{fig:fallback-entropy-fast}, although fallback steps account for only approximately one eighth of committed tokens, their tokens contribute nearly half of the total token entropy. This suggests that rather than abandoning confidence-based parallel decoding altogether, it may suffice to treat these comparatively rare, high-uncertainty fallback steps differently.

\def\FBtok{13}     %
\def\FBent{45}     %
\def\Smax{2.57}    %

\pgfmathsetmacro{\RESTtok}{100-\FBtok}
\pgfmathsetmacro{\RESTent}{100-\FBent}
\pgfmathsetmacro{\sF}{\Smax*sqrt(\FBent/100)}
\pgfmathsetmacro{\sR}{\Smax*sqrt(\RESTent/100)}
\pgfmathsetmacro{\ratio}{(\FBent/\FBtok)/(\RESTent/\RESTtok)}

\def\nohyph{\hyphenpenalty=10000 \exhyphenpenalty=10000 \emergencystretch=2em}

\definecolor{fbFork}{HTML}{D95F02}   %
\definecolor{fbRest}{HTML}{2E6FB7}   %

\begin{wrapfigure}[17]{r}{0.5\textwidth}
\vspace{-\intextsep}   %
\noindent
\begin{tikzpicture}[
    tokc/.style={circle, draw=none, minimum size=4.2mm, inner sep=0pt},
    forktok/.style={tokc, fill=fbFork},
    resttok/.style={tokc, fill=fbRest},
    rowlab/.style={font=\footnotesize\bfseries, anchor=west, inner sep=0pt},
    rowmsg/.style={font=\footnotesize\itshape\nohyph, anchor=north west,
                   text width=2.3cm, align=left, inner sep=0pt},
]
  \def\dx{0.52}      %
  \def\yrow{3.05}    %
  \def\ysq{1.05}     %
  \def\xlab{-2.60}   %
  \def\xmid{0.64}    %

  \node[anchor=center, font=\footnotesize\bfseries] at (\xmid,4.05) {%
    \tikz[baseline=-0.6ex]\fill[fbFork] (0,0) circle (1.5mm);\,Fallback Tokens%
    \hspace{0.40cm}%
    \tikz[baseline=-0.6ex]\fill[fbRest] (0,0) circle (1.5mm);\,Parallel Tokens};

  \node[rowlab] at (\xlab,{\yrow+0.20})  {Frequency};
  \node[rowmsg] at (\xlab,{\yrow-0.02})
    {Approximately 1 out of 8 tokens is a fallback token.};
  \node[rowlab] at (\xlab,{\ysq+0.20}) {Entropy};
  \node[rowmsg] at (\xlab,{\ysq-0.02})
    {Fallback tokens contribute nearly half of the total entropy.};

  \foreach \i in {0,...,7} { \node[resttok] at ({0.35+\dx*\i},\yrow) {}; }
  \node[forktok] at ({0.35+\dx*0},\yrow) {};

  \fill[fbFork] ({0.92-\sF/2},{\ysq-\sF/2}) rectangle ++(\sF,\sF);
  \fill[fbRest] ({3.03-\sR/2},{\ysq-\sR/2}) rectangle ++(\sR,\sR);

  \node[white, font=\small\bfseries] at (0.92,\ysq)
    {\pgfmathprintnumber[precision=1]{\FBent}\%};
  \node[white, font=\small\bfseries] at (3.03,\ysq)
    {\pgfmathprintnumber[precision=1]{\RESTent}\%};

\end{tikzpicture}
\caption{Token frequencies and entropies from Fast-dLLM sampling on HumanEval using LLaDA-8B-Instruct.}
\label{fig:fallback-entropy-fast}
\vspace{\baselineskip}   %
\end{wrapfigure}
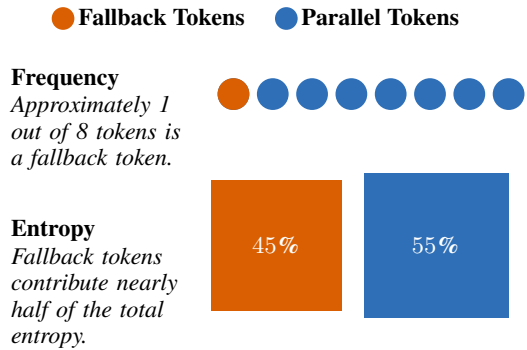

We therefore propose Fork-dLLM, which changes only the Fast-dLLM fallback, as formalized in \Cref{alg:fork-dllm}. Whenever at least one position exceeds $\lambda$, it behaves identically to Fast-dLLM and commits all such positions in parallel. When none do, Fork-dLLM instead performs an AR-style fallback by committing the leftmost remaining masked position rather than the highest-confidence one. It thus confronts uncertain decisions using AR ordering while retaining confidence-based parallel decoding elsewhere. As a one-line modification to Fast-dLLM, it requires no training, additional model evaluations, or new hyperparameters. In our experiments, Fork-dLLM matches the pass@$k$ scaling of AR sampling while requiring $2$--$3\times$ fewer NFEs (\Cref{sec:sampling}). Fast-dLLM's highest-confidence fallback is shared by several widely used confidence-based samplers
\citep{ben-hamuAcceleratedSamplingMasked2025,jazbec2025learning}, and the Fork-dLLM principle could be applied to each of them by replacing it with an AR-style one. %

\subsection{ForkGRPO}
\label{sec:methods-forkgrpo}

Next, we propose to use Fork-dLLM to make JustGRPO \citep{ni2026flexibility} more efficient in both rollout generation and policy optimization. Recall that JustGRPO generates trajectories autoregressively and evaluates the GRPO objective at every output position $k=1,\ldots,L$ (\Cref{eq:justgrpo}).
Instead, we propose to generate rollouts with Fork-dLLM (\Cref{alg:fork-dllm}) and evaluate the GRPO objective only at the fallback steps of each rollout. We call the resulting training algorithm \emph{ForkGRPO}. Since Fork-dLLM commits a single token at each fallback step, the policy-likelihood ratios at these steps are exact, without requiring the likelihood approximations used by diffu-GRPO \citep{zhao2025d}.

Concretely, for each prompt $c\sim\mathcal{D}$, ForkGRPO uses Fork-dLLM to sample a group of $G$ trajectories $\{\bm{x}_{i,T_i:0}\}_{i=1}^G$, where $T_i$ denotes the (trajectory-dependent) number of sampling steps for rollout $i$. Let $\mathcal{F}_i\subseteq[T_i]$ denote its set of \emph{fallback steps}, i.e., steps at which no masked position exceeds the confidence threshold $\lambda$:
\begin{equation*}
\mathcal{F}_i
=
\left\{
t :
\max_{k\in\mathcal{M}_{i,t}} c_{i,t}^k
\leq \lambda
\right\}.
\end{equation*}

ForkGRPO then replaces the per-position average in JustGRPO with an average only over the fallback steps determined by the Fork-dLLM sampler:
\begin{equation}
\mathcal{J}_{\mathrm{ForkGRPO}}(\theta)
=
\underset{
\substack{
c\sim\mathcal{D},\\
\raisebox{-4pt}[\height]{\makebox[0.55\width]{\colorbox{cyan!15}{$\scriptstyle
\{\bm{x}_{i,T_i:0}\}_{i=1}^G
\sim q_{\theta_{\mathrm{old}}}^{\mathrm{Fork}}(\cdot\mid c)
$}}}
}
}{\E}
\bigg[
\frac{1}{G}
\sum_{i=1}^{G}
\colorbox{cyan!15}{$
\displaystyle
\frac{1}{|\mathcal{F}_i|}
\sum_{t\in\mathcal{F}_i}
$}
\min\!\left(
\rho_{i,t}\hat A_i,\,
\operatorname{clip}(\rho_{i,t},1-\epsilon,1+\epsilon)\hat A_i
\right)
\bigg],
\label{eq:forkgrpo}
\end{equation}
where $q_{\theta_{\mathrm{old}}}^{\mathrm{Fork}}$ denotes the trajectory distribution induced by Fork-dLLM. At each fallback step $t\in\mathcal{F}_i$, let $k_{i,t}:=\operatorname{leftmost}(\mathcal{M}_{i,t})$ denote the position committed by the AR-style fallback for the $i$-th rollout. The corresponding policy-likelihood ratio is
\begin{equation*}
\rho_{i,t}
=
\frac{
q_\theta^{k_{i,t}}
\bigl(x_{i,0}^{k_{i,t}}\mid c,\bm{x}_{i,t}\bigr)
}{
q_{\theta_{\mathrm{old}}}^{k_{i,t}}
\bigl(x_{i,0}^{k_{i,t}}\mid c,\bm{x}_{i,t}\bigr)
}.
\label{eq:forkgrpo-ratio}
\end{equation*}
Observe that non-fallback tokens still affect the final rollout reward $r(\bm{x}_{i,0})$ and the states conditioning subsequent fallback decisions, but do not directly enter the policy objective. Consequently, ForkGRPO evaluates policy likelihoods only at the sparse fallback steps, which, together with its parallel rollouts, substantially reduces training cost compared to JustGRPO (\Cref{sec:grpo}).

\citet{ni2026flexibility} also propose \emph{JustGRPO-Fast}, an efficient variant of JustGRPO that evaluates the policy-likelihood ratios only at the top-$25\%$ highest-entropy positions of each rollout. Like ForkGRPO, it restricts policy optimization to a subset of positions, but it differs in two ways: it still generates rollouts autoregressively, and it selects a fixed fraction of positions by entropy. ForkGRPO instead uses parallel Fork-dLLM rollouts, which also reduces the cost of rollout generation, and selects a variable, trajectory-dependent set of positions through the fallback condition. We show experimentally that ForkGRPO substantially outperforms JustGRPO-Fast under a matched post-training budget (\Cref{fig:grpo}).

\section{Experiments}

In our experiments, we first study Fork-dLLM as a sampler, asking whether it recovers the pass@$k$ scaling of AR sampling while retaining the efficiency of parallel decoding (\Cref{sec:sampling}). We then analyze its fallback and parallel steps to better understand where these gains come from. Finally, we turn to GRPO post-training (\Cref{sec:grpo}), comparing ForkGRPO with JustGRPO-Fast \citep{ni2026flexibility} under the same compute budget and examining the entropy collapse induced by Fast-dLLM rollouts.

\subsection{Pass@$k$ Scaling}
\label{sec:sampling}

\paragraph{Setup.} We evaluate on two popular masked dLLMs, {LLaDA-8B-Instruct} \citep{nie2026large} and {Dream-v0-Instruct-7B} \citep{ye2025dream}, across four standard benchmarks in mathematical reasoning and code generation: {GSM8K} \citep{cobbe2021gsm8k}, {MATH-500} \citep{hendrycksmath2021}, {HumanEval} \citep{chen2021evaluatinglargelanguagemodels}, and {MBPP} \citep{austin2021programsynthesislargelanguage}.\footnote{Following prior work, we evaluate pass@$k$ scaling on the first $N_{test}=300$ test samples of GSM8K to reduce computational costs \citep{olausson2026tale}. Additionally, for MBPP we exclude the longest prompt ($i=493$), a significant outlier in length.}
We use the common semi-AR regime with block length $BL=32$ and generation length $L=256$.

We report pass@$k$, the standard measure of whether at least one of $k$ samples solves a given problem \citep{kulal2019spoc, chen2021evaluatinglargelanguagemodels, ni2026flexibility}. We additionally report the cost-adjusted pass@NFE, which measures pass rate as a function of inference compute and avoids favoring methods that use more NFEs per sample. For AR sampling, we count one NFE per token generated before the \texttt{〈eos〉} token.

Alongside Fast-dLLM and AR sampling, we compare Fork-dLLM against Tempered Confidence Thresholding (TCT) \citep{olausson2026tale}, a recent approach that mitigates the flexibility trap using a position-temperature hyperparameter to randomize which positions are unmasked. For Fork-dLLM, Fast-dLLM, and TCT, we sweep the relevant sampling temperatures and confidence threshold $\lambda$ as described in \Cref{app:sampling-hparams}, and report the best-performing configuration for each method. Following prior work, we use a sampling temperature of $\tau=0.6$ for AR generation \citep{ni2026flexibility,olausson2026tale}.

\begin{figure}[t]
    \centering
    \includegraphics[width=\textwidth]{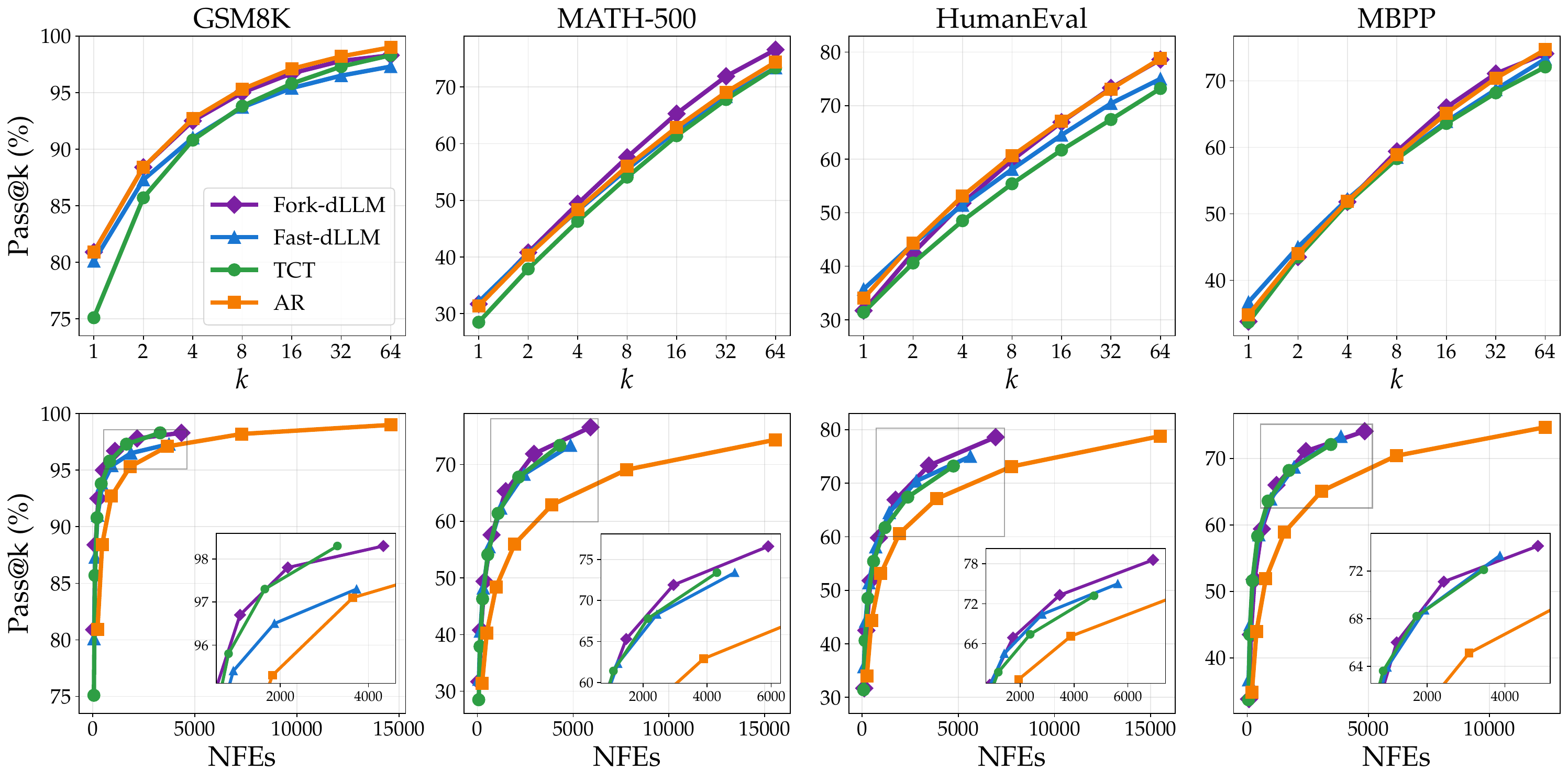}
    \caption{
Pass@$k$ (top) and pass@NFE (bottom) curves on LLaDA-8B-Instruct across mathematics and code benchmarks. Both rows show the same generations; the bottom row instead accounts for the number of NFEs required to obtain the corresponding samples. Fork-dLLM, together with the other two confidence-based samplers (Fast-dLLM and TCT), retains the strong pass@$k$ scaling of AR sampling while requiring substantially fewer NFEs to reach a given performance level. Insets zoom into the high-accuracy regime.
}
    \label{fig:passk-llada}
\end{figure}

\paragraph{Pass@$k$ scaling.}

We begin with the pass@$k$ and pass@NFE scaling of Fork-dLLM against the baselines. \Cref{fig:passk-llada} gives the LLaDA-8B-Instruct results, with Dream-v0-Instruct-7B in \Cref{fig:passk-dream}. We observe that all confidence-based samplers exhibit strong pass@$k$ scaling, remaining surprisingly close to AR across datasets. In particular, Fast-dLLM performs substantially better than suggested by prior comparisons \citep{olausson2026tale}, which did not jointly tune these hyperparameters. Fork-dLLM is generally the strongest of the parallel samplers, particularly at larger $k$ (see insets in \Cref{fig:passk-llada}). Importantly, when accounting for inference cost, all three confidence-based methods substantially outperform AR in pass@NFE, demonstrating that strong pass@$k$ scaling does not require giving up the efficiency of parallel decoding.

\paragraph{Analysing fallback and parallel steps.} Prior work identifies \emph{fork} tokens as high-entropy branching points that determine subsequent reasoning directions \citep{wang2026beyond}, and shows that arbitrary-order generation frequently bypasses them \citep{ni2026flexibility}. If Fork-dLLM's fallback steps capture these branching points, the tokens committed during fallback should therefore resemble fork tokens. To examine this, \Cref{fig:token-clouds} shows the tokens Fork-dLLM commits most often during parallel and fallback steps. Indeed, fallback steps are dominated by connectives such as \textit{Since}, \textit{First}, and \textit{Thus}, consistent with Fork-dLLM resolving forks directly rather than deferring them. In contrast, parallel steps predominantly commit dataset-specific words that follow an already established reasoning path. Both clouds closely resemble those reported by \cite{wang2026beyond} for high- and low-entropy tokens in AR models.

\definecolor{forkI}{HTML}{8C2D04}
\definecolor{forkII}{HTML}{D95F02}
\definecolor{forkIII}{HTML}{EFA033}
\definecolor{forkIV}{HTML}{A6432A}
\definecolor{forkV}{HTML}{C7761B}
\definecolor{parI}{HTML}{1B3A6B}
\definecolor{parII}{HTML}{2E6FB7}
\definecolor{parIII}{HTML}{4C9BD1}
\definecolor{parIV}{HTML}{5F4B8B}
\definecolor{parV}{HTML}{2A7F7F}

\providecommand{\wcmin}{7}
\providecommand{\wcmax}{14}
\providecommand{\wclo}{1}
\providecommand{\wchi}{1}
\providecommand{\wcfont}{\ttfamily}
\providecommand{\wcw}[5]{%
  \pgfmathsetmacro{\wcsize}{\wcmin+(\wcmax-\wcmin)*ln(#1/\wclo)/ln(\wchi/\wclo)}%
  \node[text=#2, inner sep=0pt] at (#3,#4)
    {\fontsize{\wcsize}{\wcsize}\selectfont\wcfont #5};%
}

\begin{figure}[t]
\centering
\begin{subfigure}[t]{0.48\textwidth}
\centering
\begin{tikzpicture}
  \renewcommand{\wclo}{0.16}\renewcommand{\wchi}{1.29}
  \useasboundingbox (0,0) rectangle (6.60,2.39);
  \wcw{0.23}{forkI}{0.49}{2.23}{There}
  \wcw{0.23}{forkII}{1.89}{2.23}{consider}
  \wcw{0.17}{forkIII}{3.46}{2.23}{multiply}
  \wcw{0.24}{forkIV}{4.69}{2.23}{this}
  \wcw{0.56}{forkV}{5.54}{2.23}{To}
  \wcw{0.20}{forkI}{6.29}{2.23}{She}
  \wcw{0.30}{forkII}{0.35}{1.81}{The}
  \wcw{0.17}{forkIII}{1.58}{1.81}{initially}
  \wcw{0.20}{forkIV}{2.93}{1.81}{gives}
  \wcw{0.43}{forkV}{3.92}{1.81}{Let}
  \wcw{0.50}{forkI}{4.74}{1.81}{We}
  \wcw{0.44}{forkII}{5.88}{1.81}{follow}
  \wcw{0.30}{forkIII}{0.35}{1.36}{Now}
  \wcw{0.17}{forkIV}{1.08}{1.36}{He}
  \wcw{1.01}{forkV}{2.21}{1.36}{Since}
  \wcw{0.23}{forkI}{3.80}{1.36}{Finally}
  \wcw{0.21}{forkII}{5.11}{1.36}{After}
  \wcw{0.61}{forkIII}{6.18}{1.36}{let}
  \wcw{0.23}{forkIV}{0.58}{0.95}{Adding}
  \wcw{0.26}{forkV}{1.74}{0.95}{Thus}
  \wcw{0.16}{forkI}{2.83}{0.95}{needed}
  \wcw{0.20}{forkII}{4.29}{0.95}{Determine}
  \wcw{0.17}{forkIII}{5.60}{0.95}{over}
  \wcw{0.24}{forkIV}{6.36}{0.95}{So}
  \wcw{1.29}{forkV}{0.80}{0.52}{First}
  \wcw{0.16}{forkI}{2.27}{0.52}{annual}
  \wcw{0.17}{forkII}{3.38}{0.52}{given}
  \wcw{0.17}{forkIII}{4.20}{0.52}{On}
  \wcw{0.16}{forkIV}{4.93}{0.52}{This}
  \wcw{0.36}{forkV}{6.03}{0.52}{there}
\end{tikzpicture}
\caption{Frequent fallback tokens}
\label{fig:fork-cloud}
\end{subfigure}
\hfill
\begin{subfigure}[t]{0.48\textwidth}
\centering
\begin{tikzpicture}
  \renewcommand{\wclo}{0.07}\renewcommand{\wchi}{0.72}
  \useasboundingbox (0,0) rectangle (6.60,2.39);
  \wcw{0.08}{parI}{0.53}{2.34}{pounds}
  \wcw{0.08}{parII}{1.67}{2.34}{costs}
  \wcw{0.14}{parIII}{2.82}{2.34}{years}
  \wcw{0.46}{parIV}{3.72}{2.34}{7}
  \wcw{0.09}{parV}{4.46}{2.34}{days}
  \wcw{0.70}{parI}{5.81}{2.34}{times}
  \wcw{0.20}{parII}{0.49}{1.88}{frac}
  \wcw{0.09}{parIII}{1.41}{1.88}{age}
  \wcw{0.14}{parIV}{2.29}{1.88}{than}
  \wcw{0.31}{parV}{3.64}{1.88}{reason}
  \wcw{0.24}{parI}{5.18}{1.88}{hours}
  \wcw{0.12}{parII}{6.27}{1.88}{day}
  \wcw{0.08}{parIII}{0.37}{1.41}{year}
  \wcw{0.72}{parIV}{1.85}{1.41}{answer}
  \wcw{0.33}{parV}{3.14}{1.41}{9}
  \wcw{0.11}{parI}{3.90}{1.41}{time}
  \wcw{0.08}{parII}{5.01}{1.41}{second}
  \wcw{0.15}{parIII}{6.15}{1.41}{week}
  \wcw{0.21}{parIV}{0.49}{0.91}{many}
  \wcw{0.70}{parV}{1.35}{0.91}{8}
  \wcw{0.07}{parI}{2.08}{0.91}{hour}
  \wcw{0.72}{parII}{3.24}{0.91}{text}
  \wcw{0.15}{parIII}{4.80}{0.91}{dollars}
  \wcw{0.08}{parIV}{6.15}{0.91}{steps}
  \wcw{0.13}{parV}{0.71}{0.42}{minutes}
  \wcw{0.09}{parI}{2.07}{0.42}{water}
  \wcw{0.33}{parII}{3.15}{0.42}{ing}
  \wcw{0.07}{parIII}{3.89}{0.42}{x}
  \wcw{0.09}{parIV}{4.59}{0.42}{more}
  \wcw{0.45}{parV}{5.88}{0.42}{boxed}
\end{tikzpicture}
\caption{Frequent parallel tokens}
\label{fig:parallel-cloud}
\end{subfigure}
\caption{The 30 most frequent tokens in fallback steps (a) and parallel steps (b) from Fork-dLLM on MATH-500 using LLaDA-8B-Instruct. Word font size is scaled to frequency.%
}
\label{fig:token-clouds}
\vspace{-10pt}
\end{figure}

\begin{wrapfigure}{r}{0.55\textwidth}
\vspace{-0.75\intextsep}   %
    \centering
    \includegraphics[width=\linewidth]{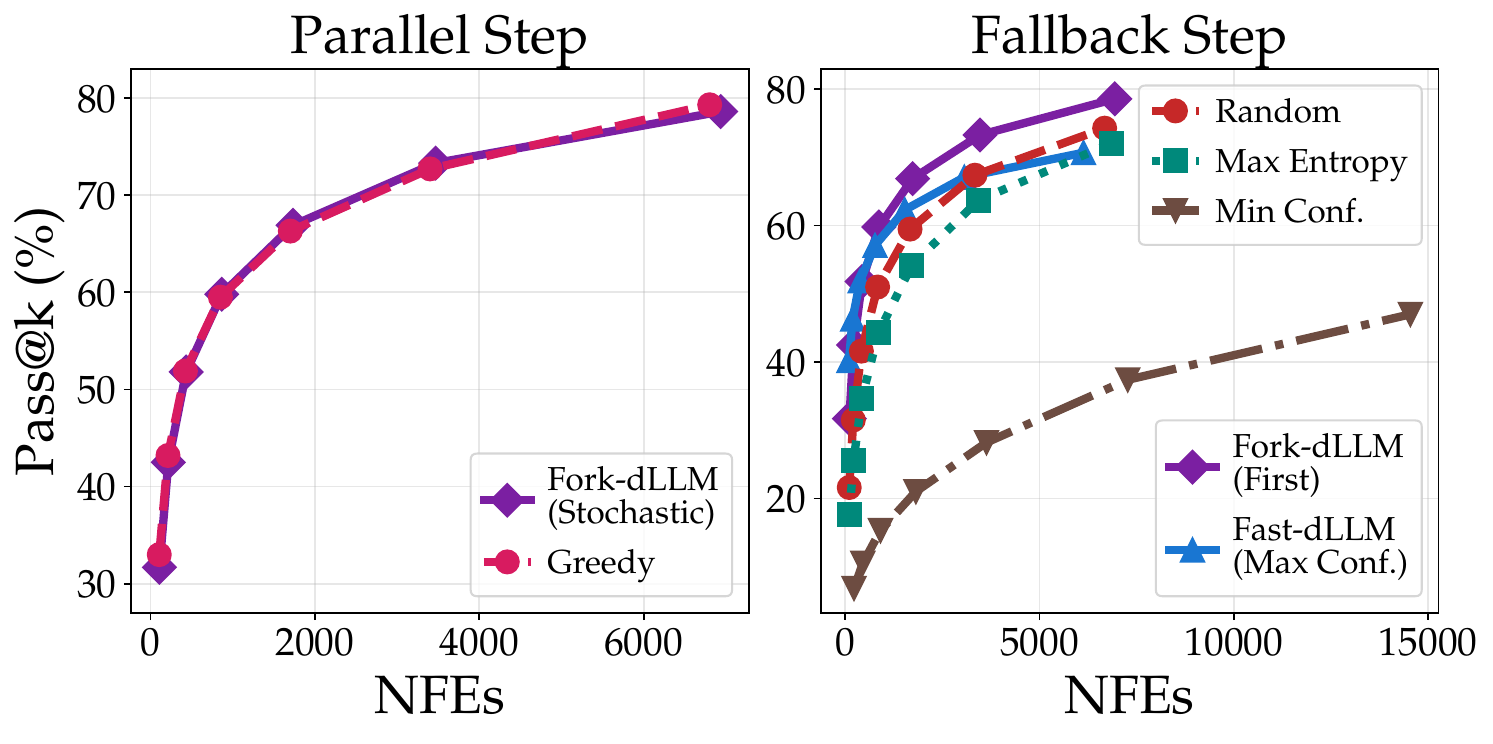}
    \caption{Pass@NFE curves of Fork-dLLM and its ablations on HumanEval using LLaDA-8B-Instruct.}
    \label{fig:fork-ablations-nfe}
\vspace{-\baselineskip}   %
\end{wrapfigure}

To identify which component drives Fork-dLLM's diversity gains, we separately ablate the parallel and fallback steps. \Cref{fig:fork-ablations-nfe} reports pass@NFE, with the corresponding pass@$k$ curves in \Cref{fig:fork-ablations-k}. Making parallel generation greedy ($\tau=0$) instead of stochastic leaves the scaling essentially unchanged (\emph{left}). In contrast, replacing Fork-dLLM's AR-style fallback with lowest-confidence, highest-entropy, or random position selection leads to worse scaling (\emph{right}). Together, these results indicate that the fallback ordering, rather than stochasticity in the parallel steps, drives most of Fork-dLLM's gains.

\subsection{GRPO Training}
\label{sec:grpo}

We next evaluate whether the benefits of Fork-dLLM carry over to GRPO post-training. We compare ForkGRPO against AR-based and parallel-decoding baselines under a fixed training compute budget, focusing on both training efficiency and downstream performance.

\paragraph{Setup.} We compare our proposed ForkGRPO against three baselines: FastGRPO, JustGRPO, and JustGRPO-Fast. FastGRPO is a parallel-decoding baseline we introduce, which, like ForkGRPO, applies policy updates only at fallback steps but generates rollouts with Fast-dLLM. Thus, the two methods differ in which positions trigger these updates: FastGRPO inherits Fast-dLLM's confidence-based fallback, whereas ForkGRPO uses Fork-dLLM's AR-style fallback. JustGRPO and JustGRPO-Fast \citep{ni2026flexibility}  instead use autoregressive rollouts, with JustGRPO-Fast restricting the objective to the highest-entropy positions. Following \citet{wang2026beyond}, we set this fraction to $20\%$. All variants use generation length $L=256$, while ForkGRPO and FastGRPO additionally use block length $BL=32$. We take the sampling temperatures and confidence thresholds from \Cref{app:sampling-hparams}.

We follow the standard GRPO formulation with group-standardized advantages, using a group size of $G=16$ and $16$ prompts per step. For mathematics, an answer receives a reward of $1$ if it is correct and $0$ otherwise; for code, the reward is the fraction of unit tests passed. We use no KL penalty ($\beta=0$). All runs start from LLaDA-8B-Instruct \citep{nie2026large}. Unlike \cite{ni2026flexibility}, who fine-tune the full model, we train LoRA adapters ($r=128$, $\alpha=64$) on the attention and MLP projections and keep the base weights frozen.

We train ForkGRPO, FastGRPO and JustGRPO-Fast ourselves. Due to the substantially higher computational cost of full JustGRPO training, we use the released JustGRPO checkpoint\footnote{\url{https://github.com/LeapLabTHU/JustGRPO}} from \cite{ni2026flexibility}, together with its reported training configuration. For mathematics, we train and evaluate on the training and test splits of GSM8K and MATH-500.
For code generation, we train on the subset of AceCoder \citep{zeng2025acecoder} selected by \cite{gong2026diffucoder} and evaluate on HumanEval and MBPP.
Each of our runs is given a fixed compute budget of $4\times$H100 GPUs for $12$ hours. We evaluate with greedy Fast-dLLM decoding, sweeping the confidence threshold, over the full test sets. \Cref{tab:grpo-hparams} lists the full configuration.

\begin{figure}[t]
    \centering
    \includegraphics[width=\linewidth]{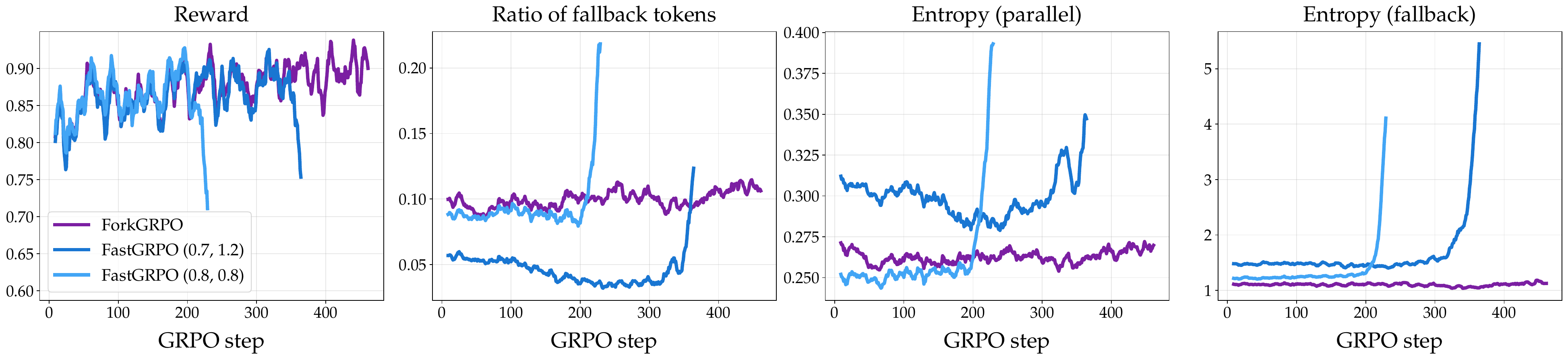}
    \caption{Reward, ratio of fallback tokens, and entropy of the tokens committed in parallel and fallback steps over GRPO training on GSM8K. %
    FastGRPO is shown with the tuned Fast-dLLM values, $(\lambda, \tau) = (0.7, 1.2)$, and with those of ForkGRPO, $(0.8, 0.8)$. In an attempt to improve stability, this experiment adds a KL penalty of $\beta=0.02$; we observe similar behaviour at $\beta=0$.}
    \label{fig:grpo-fast-collapse}
    \vspace{-10pt}
\end{figure}

\paragraph{FastGRPO collapse.} Before turning to AR rollouts, we compare ForkGRPO with FastGRPO. We train FastGRPO with the tuned Fast-dLLM values ($\lambda=0.7$, $\tau=1.2$) and with the ForkGRPO values ($\lambda=0.8$, $\tau=0.8$); with the latter, the two methods differ only in how the fallback position is selected. \Cref{fig:grpo-fast-collapse} tracks all three runs during GSM8K training. Both FastGRPO runs initially match ForkGRPO in reward but collapse after roughly $350$ and $220$ steps, respectively. Each collapse coincides with a sharp rise in the fallback ratio and in the entropy of parallel and fallback tokens, with fallback entropy growing from below $1.5$ to above $4$. With the tuned values, the collapse is also preceded by a steady decline of the fallback ratio from $0.06$ to below $0.04$, as the sampler commits more positions in parallel. An extended $24$-hour ForkGRPO run shows none of this: all three quantities remain stable and the reward does not collapse. This is despite Fast-dLLM and Fork-dLLM showing similar pass@$k$ scaling at inference (\Cref{fig:passk-llada}). We hypothesize that selecting the fallback position independently of model confidence decouples the rollout policy from RL-induced changes in confidence, stabilizing training. Understanding this mechanism more fully warrants further investigation.

\paragraph{ForkGRPO vs.\ JustGRPO-Fast under matched compute.}

\begin{figure}[t]
    \centering
    \includegraphics[width=\linewidth]{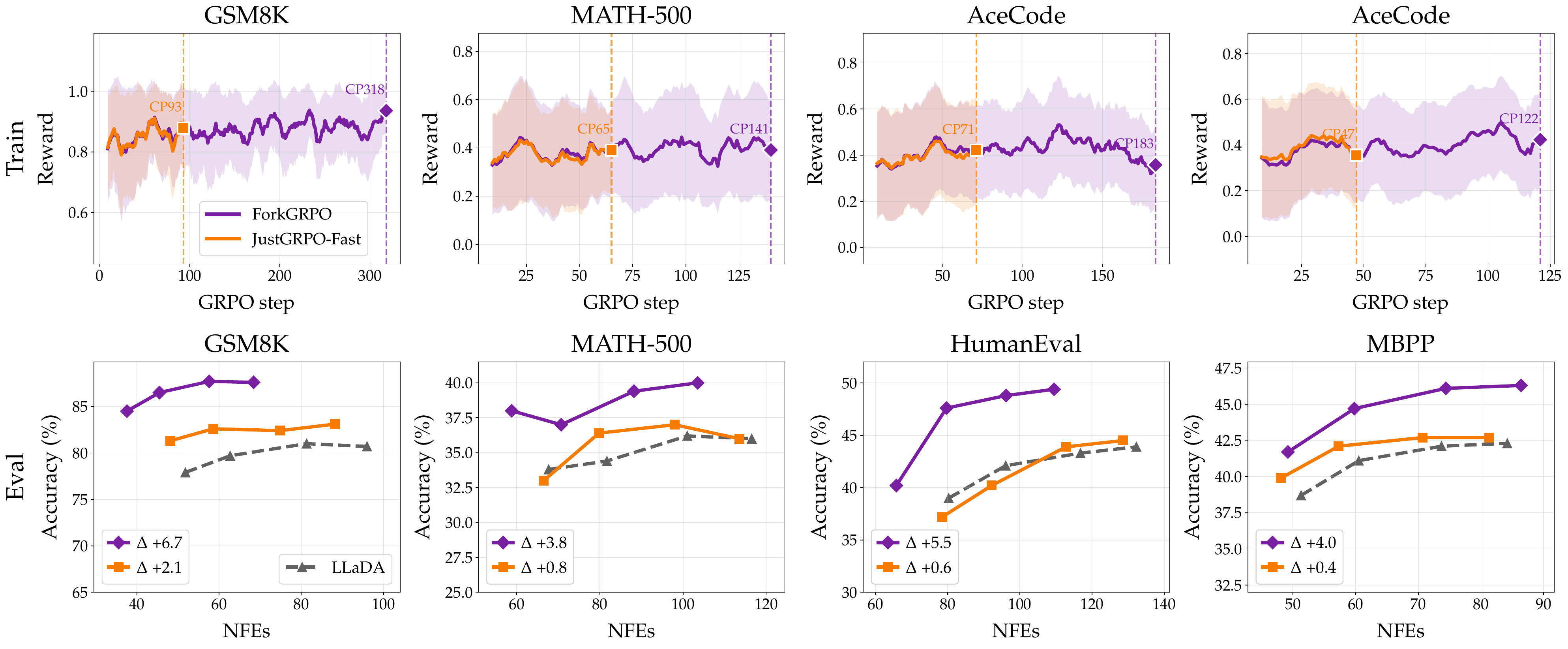}
    \caption{Training reward (top) and final-checkpoint accuracy versus NFEs (bottom) for ForkGRPO and JustGRPO-Fast under the same training budget ($4\times$H100, 12\,h). Dashed vertical lines in the top row mark the final GRPO step reached by each method: owing to cheaper rollouts and policy optimization, ForkGRPO completes more than twice as many training steps. The bottom row evaluates the resulting checkpoints with Fast-dLLM decoding ($\tau=0$), sweeping the confidence threshold $\lambda$. ForkGRPO yields consistently larger improvements over the base model than JustGRPO-Fast.}
    \label{fig:grpo}
\end{figure}

We next compare ForkGRPO with JustGRPO-Fast on mathematics and code benchmarks under the same compute budget (\Cref{fig:grpo}). On every benchmark, ForkGRPO completes more than twice as many training steps, for example $3.4\times$ as many on GSM8K. It also improves consistently over LLaDA-8B-Instruct, whereas the gains of JustGRPO-Fast remain small ($+6.7$ vs.\ $+2.1$ points on GSM8K, $+5.5$ vs.\ $+0.6$ on HumanEval). On GSM8K, MATH-500, and HumanEval, ForkGRPO checkpoints also decode with fewer NFEs at every threshold.

Since part of ForkGRPO's speed-up comes from parallel rollouts, we additionally compare checkpoints of the two methods at matched training steps (\Cref{tab:grpo-step-matched}). The two methods reach similar accuracy, but JustGRPO-Fast requires $1.6$--$1.9\times$ more backward passes, as it trains on more positions per answer. This suggests that Fork-dLLM's fallback positions are more informative per update than the entropy-percentile selection of JustGRPO-Fast. These results extend those of \cite{olausson2026tale}, who train with their proposed TCT sampler and the d1 policy-likelihood approximation \citep{zhao2025d}: under a matched compute budget, AR rollouts are not necessary for effective post-training of dLLMs.

\begin{table}[t]
    \centering
    \begin{tabular}{llcccc}
        \toprule
        & & \multicolumn{2}{c}{ForkGRPO} & \multicolumn{2}{c}{JustGRPO-Fast \textit{(relative)}} \\
        \cmidrule(lr){3-4}\cmidrule(lr){5-6}
        Benchmark & Step & Acc. & Bwd. passes & Acc. ($\Delta$) & Bwd. passes ($\times$) \\
        \midrule
        \multirow{2}{*}{GSM8K}    & $50$  & $82.9$ & $3.2$k & $-0.6$ & $1.72\times$ \\
                                  & $100$ & $83.8$ & $5.3$k & $+0.1$ & $1.92\times$ \\
        \midrule
        \multirow{2}{*}{MATH-500} & $25$  & $35.8$ & $4.2$k & $-0.2$ & $1.60\times$ \\
                                  & $50$  & $37.0$ & $8.0$k & $-1.6$ & $1.63\times$ \\
        \bottomrule
    \end{tabular}
    \caption{Step-matched accuracy and backward passes, evaluated at $\lambda=0.95$. JustGRPO-Fast is reported relative to ForkGRPO, with accuracy differences in percentage points. %
    }
    \label{tab:grpo-step-matched}
\end{table}

\paragraph{ForkGRPO vs.\ JustGRPO.}

Finally, we compare ForkGRPO, trained with LoRA on $4\times$H100 for $12$\,h, with the JustGRPO checkpoint released by \citet{ni2026flexibility}, trained with full fine-tuning on $16\times$H100 for $24$\,h. \Cref{fig:checkpoints-all} evaluates both checkpoints using Fast-dLLM ($\lambda\in\{0.7,0.8,0.9,0.95\}$) and the EB sampler \citep{ben-hamuAcceleratedSamplingMasked2025} ($\gamma\in\{0.1,0.5,1\}$). Despite its substantially smaller training budget, ForkGRPO reaches comparable peak accuracy on GSM8K, while JustGRPO attains similar accuracy at lower NFEs. On MATH-500, ForkGRPO performs consistently better under both samplers, reaching roughly $40\%$ accuracy compared with $37$--$38\%$ for JustGRPO.

\section{Conclusion}

We introduced Fork-dLLM, which replaces the confidence-based fallback of Fast-dLLM with an AR-style one and otherwise keeps parallel decoding unchanged. Across mathematics and code benchmarks, Fork-dLLM matches the pass@$k$ scaling of AR sampling while requiring $2$--$3\times$ fewer NFEs. We further introduced ForkGRPO, which extends this principle to RL post-training by optimizing only at fallback steps, where exact policy-likelihood ratios are available. It substantially outperforms AR-based GRPO under the same compute budget, and matches it at a fraction of the training cost.

\paragraph{Limitations and future work.}
For a fair comparison, we tuned the sampling hyperparameters of all confidence-based samplers in the same setting, HumanEval with LLaDA-8B-Instruct, and used them unchanged elsewhere; however, tuning them per model and benchmark may yield better operating points, including that of Fork-dLLM. Our GRPO experiments use a single configuration: LoRA adapters on LLaDA-8B-Instruct, trained on $4\times$H100 GPUs for $12$ hours. Testing ForkGRPO on other models, with full fine-tuning, and with longer training is left to future work, as are comparisons to other RL methods for dLLMs. Due to computational costs, we compare against the released JustGRPO checkpoint of \cite{ni2026flexibility} rather than retraining JustGRPO under our training setup. Future work should also investigate further why ForkGRPO remains stable while FastGRPO collapses during post-training (\Cref{fig:grpo-fast-collapse}), apply the AR-style fallback to other confidence-based samplers such as EB \citep{ben-hamuAcceleratedSamplingMasked2025}, and extend both Fork-dLLM and ForkGRPO to uniform discrete dLLMs \citep{chen2026dmax, team2026diffusiongemma}.

\bibliography{main}
\bibliographystyle{iclr2027_conference}

\clearpage
\appendix

\section{Related Work}
\label{sec:rel-work}

\paragraph{Diverse sampling in dLLMs.} Confidence-based samplers enable efficient parallel decoding, but recent work shows that deterministic confidence ordering can substantially reduce diversity across repeated generations \citep{ni2026flexibility,olausson2026tale}. Several training-free methods address this from different directions: TCT stochastically relaxes confidence-based ordering \citep{olausson2026tale}, TAPS perturbs early denoising states to encourage semantic branching \citep{wu2026time}, feature-space repulsion discourages redundant samples within a batch \citep{lamont2026free}, and Order-Token Search jointly explores token values and generation orders \citep{shen2026improving}. Related work also identifies low-confidence or high-entropy regions as particularly important points at which to explore \citep{fu2025bitsroundsparalleldecoding}. %
Fork-dLLM differs in that it changes only Fast-dLLM's existing fallback: when no position clears the confidence threshold, it commits the leftmost masked position, requiring no search, additional model evaluations, or new hyperparameters.

\paragraph{RL post-training for dLLMs.}
Applying reinforcement learning to dLLMs is challenging because their non-autoregressive denoising process does not provide the tractable left-to-right sequence likelihood used by standard policy optimization.
Early approaches therefore rely on approximate likelihoods or policy-gradient estimators tailored to the diffusion process \citep{zhao2025d,gong2026diffucoder,tang2026wd1,wang2026spg}.
Subsequent work has developed sequence-level ELBO objectives \citep{rojas2026improving,ou2026principled}, more stable or accurate likelihood-ratio estimators \citep{zhong2026stabilizing,monsefi2026daca}, and likelihood-free alternatives such as guided denoiser self-distillation \citep{tang2026gdsd}. Other approaches exploit the denoising trajectory explicitly, for example through trajectory-aligned training \citep{wang2026revolutionizing}, entropy-guided selection of informative optimization steps \citep{kunde2026reinforcement}, or jointly optimizing token and unmasking decisions \citep{raajesh2026mask}.
TraFL \citep{ahmadi2026beyond} additionally studies the loss of solution coverage induced by reward-maximizing post-training.
JustGRPO \citep{ni2026flexibility} takes a different approach: it switches to AR rollouts, for which the trajectory-level likelihood factorizes exactly \citep{chen2025dultra,turok2026duel}, thereby avoiding likelihood approximations at the cost of abandoning parallel generation. ForkGRPO retains parallel Fork-dLLM rollouts and applies the GRPO objective only at its comparatively sparse AR-style fallback steps, where exact token-level likelihood ratios are available.
\newpage
\section{Hyperparameters}
\label{app:hparams}

\subsection{Stochastic Sampling}
\label{app:sampling-hparams}

\begin{wraptable}{r}{0.35\textwidth}
\vspace{-\intextsep}   %
    \centering
    \begin{tabular}{lcc}
        \toprule
        Method & $\lambda$ & $\tau$ \\
        \midrule
        AR & -- & $0.6$ \\
        Fork-dLLM & $0.8$ & $0.8$ \\
        Fast-dLLM & $0.7$ & $1.2$ \\
        TCT & $0.7$ & $0.8$ \\
        \bottomrule
    \end{tabular}
    \caption{Selected hyperparameter values for each sampling method.}
    \label{tab:method-hparams}
\end{wraptable}

\begin{figure}[ht]
    \centering
    \includegraphics[width=\textwidth]{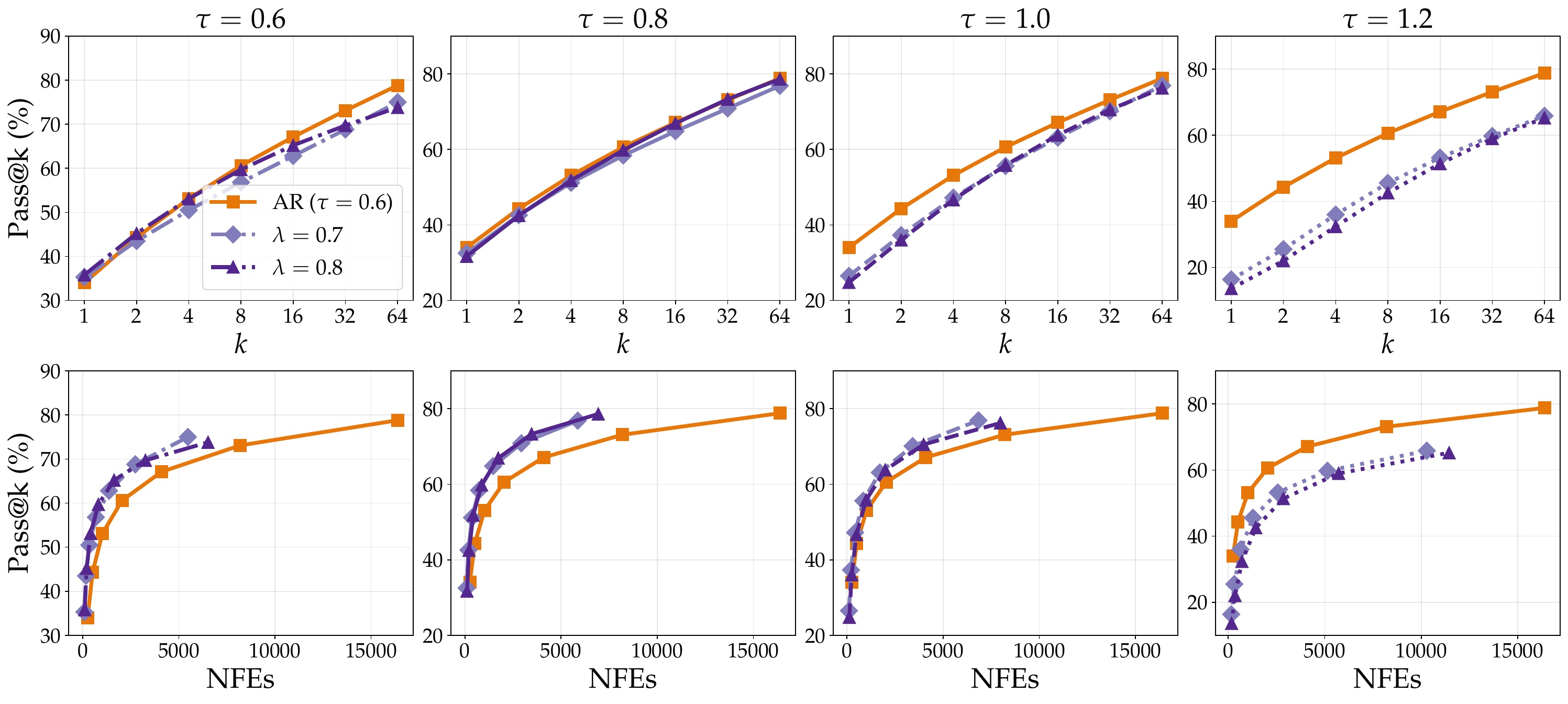}
    \caption{Pass@$k$ and pass@NFE curves for Fork-dLLM. AR is shown as a baseline.}
    \label{fig:fork-hyperparms}
\end{figure}

\begin{figure}[h]
    \centering
    \includegraphics[width=\textwidth]{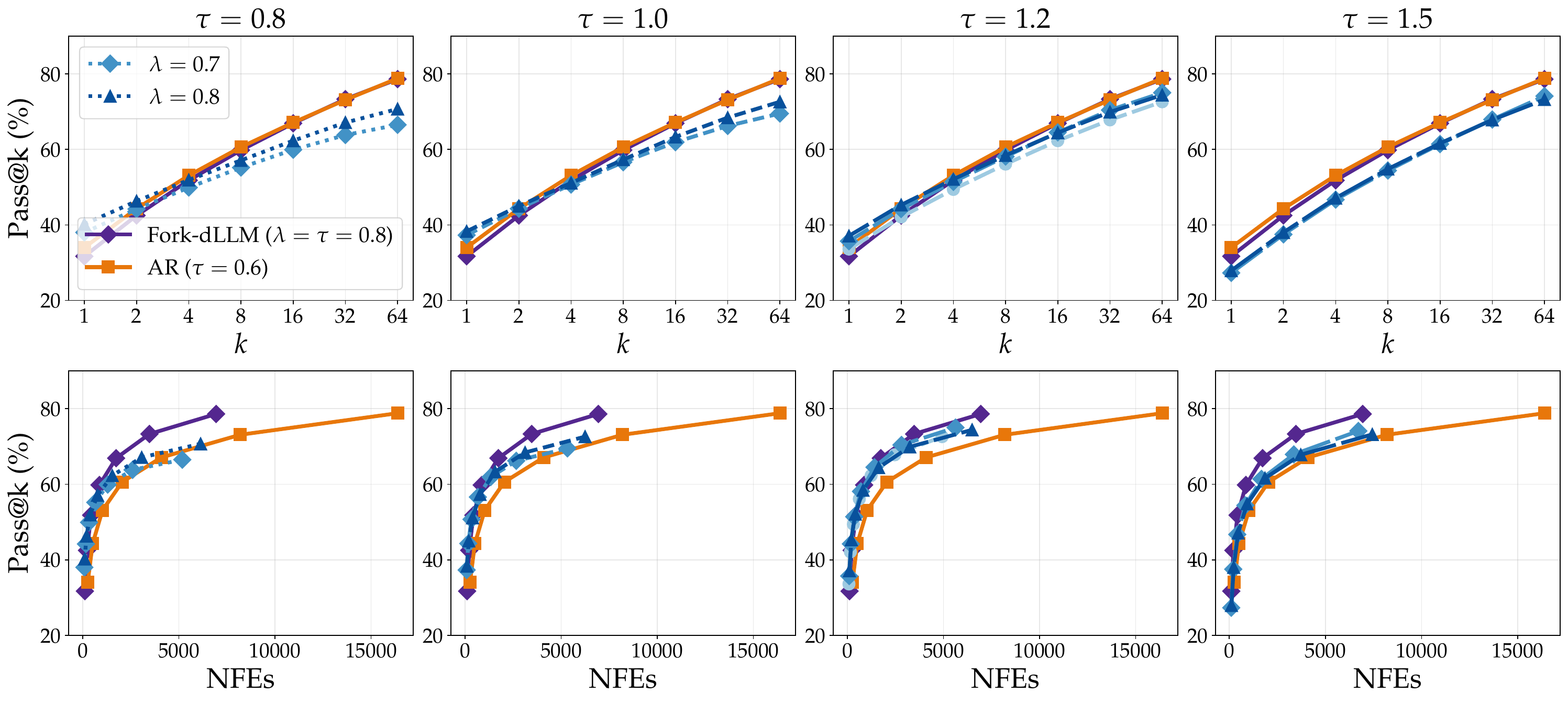}
    \caption{Pass@$k$ and pass@NFE curves for Fast-dLLM. AR and Fork-dLLM are shown as baselines.}
    \label{fig:fast-hyperparms}
\end{figure}

\begin{figure}[t]
    \centering
    \includegraphics[width=0.8\textwidth]{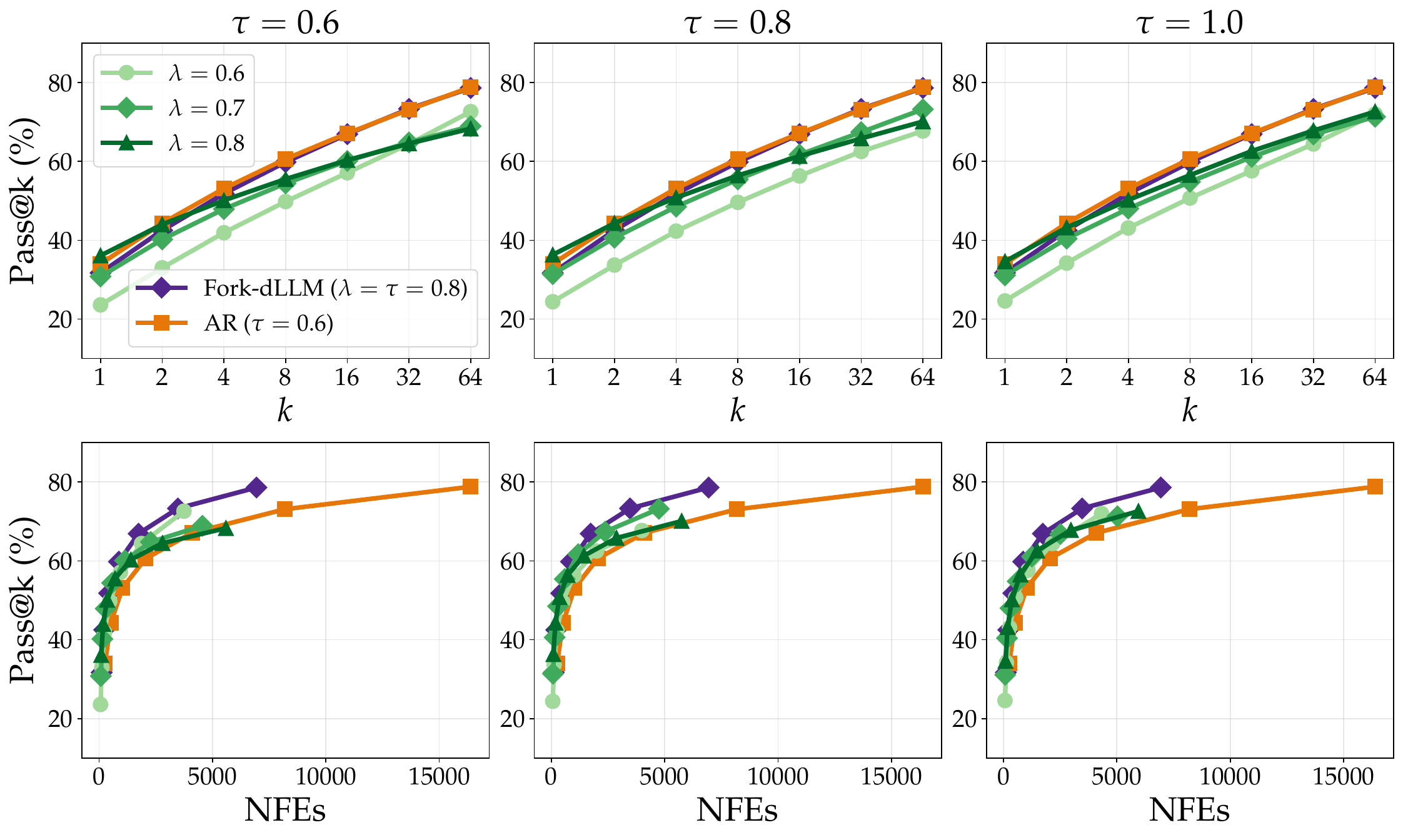}
    \caption{Pass@$k$ and pass@NFE curves for TCT. AR and Fork-dLLM are shown as baselines.}
    \label{fig:tct-hyperparms}
\end{figure}

To find the optimal hyperparameter values, we vary the confidence threshold $\lambda$ and sampling temperature $\tau$ for Fork-dLLM, Fast-dLLM, and TCT. The corresponding pass@$k$ curves are shown in Figures \ref{fig:fork-hyperparms}, \ref{fig:fast-hyperparms} and \ref{fig:tct-hyperparms}. Each figure shows pass@$k$ (top, log-scale) and pass@NFE (bottom) curves; both rows plot the same data, only the metric changes. Following \cite{olausson2026tale}, we fix TCT's additional position temperature at $\tau_{pos} = 0.1$. All experiments were run using LLaDA-8B-Instruct on HumanEval. We report the optimal values in \Cref{tab:method-hparams}.

\vspace{30pt}

\subsection{GRPO Training}
\label{app:grpo-hparams}
\Cref{tab:grpo-hparams} lists the full GRPO configuration.

\begin{table}[ht]
    \centering
    \begin{tabular}{lll}
        \toprule
        Group & Hyperparameter & Value \\
        \midrule
        \multirow{6}{*}{Objective}
            & Group Size ($G$)               & $16$ \\
            & Prompts per Step               & $16$ \\
            & Groups with $\sigma_G=0$       & Skipped \\
            & KL Penalty ($\beta$)           & $0$ \\
            & Policy Update Steps ($\mu$)    & $1$ \\
            & Top-Entropy Ratio$^*$          & $20\%$ \\
        \midrule
        \multirow{6}{*}{Optimization}
            & Optimizer                      & AdamW \\
            & Adam Betas                     & $(0.9, 0.99)$ \\
            & Learning Rate                  & $2\times10^{-5}$ \\
            & Warmup Steps                   & $0$ \\
            & Weight Decay                   & $0$ \\
            & Gradient Clipping              & $0.5$ \\
        \midrule
        \multirow{3}{*}{LoRA}
            & Rank ($r$)                     & $128$ \\
            & Alpha ($\alpha$)               & $64$ \\
            & Dropout                        & $0.05$ \\
        \midrule
        \multirow{2}{*}{Compute}
            & GPUs                           & $4\times$H100 \\
            & Training Time                  & $12$\,h \\
        \bottomrule
    \end{tabular}
    \caption{GRPO hyperparameters. The row marked $^*$ applies only to JustGRPO-Fast. With a single policy update step per rollout, the clipping range $\epsilon$ never binds and is omitted.}
    \label{tab:grpo-hparams}
\end{table}

\clearpage
\section{Additional Figures}
\label{app:figures}

\begin{figure}[ht]
    \centering
    \includegraphics[width=\textwidth]{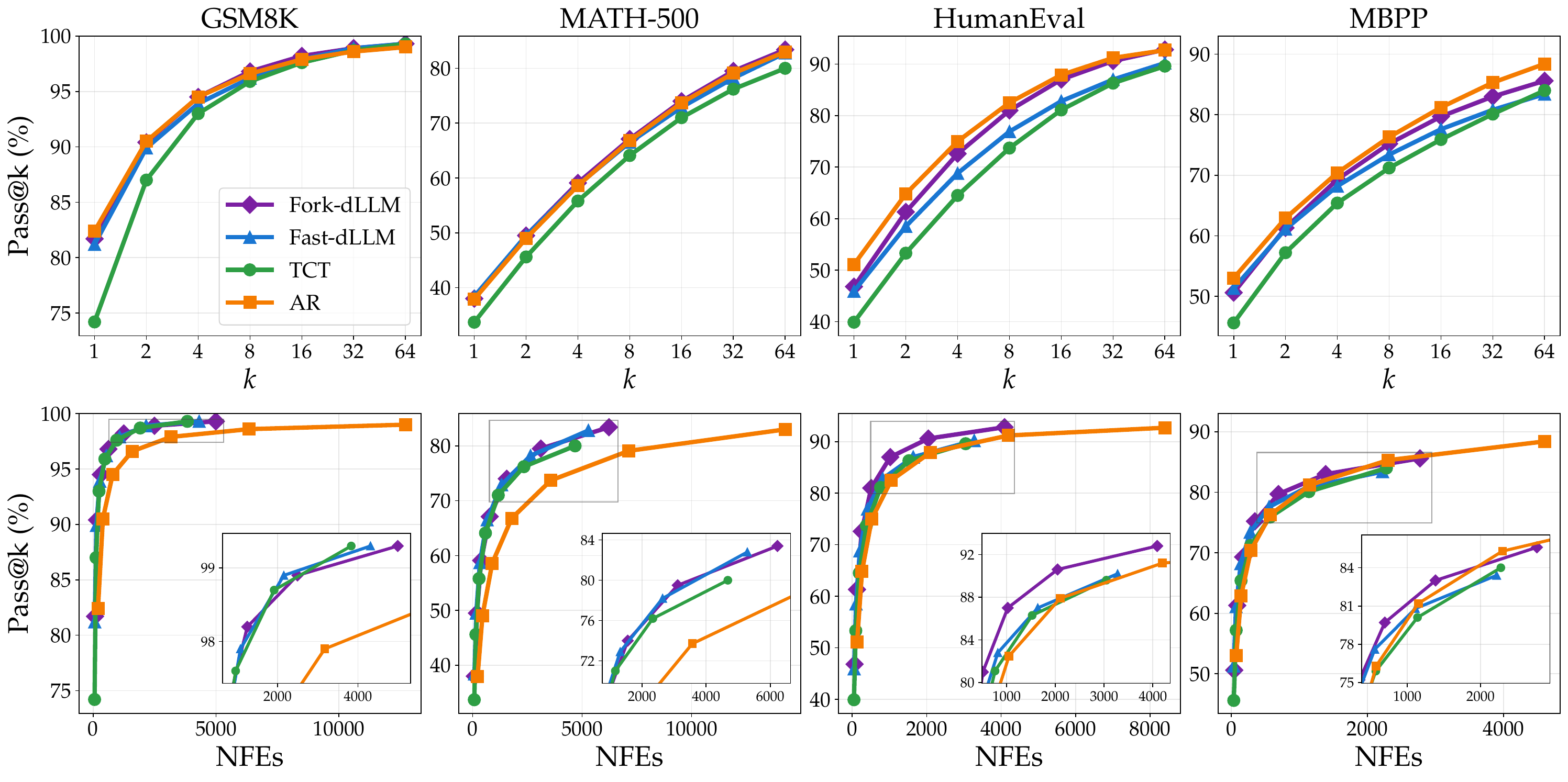}
    \caption{Pass@$k$ (top) and pass@NFE (bottom) curves on Dream-v0-Instruct-7B.}
    \label{fig:passk-dream}
\end{figure}

\Cref{fig:passk-dream} shows the results on Dream-v0-Instruct-7B. On the mathematics benchmarks, Fork-dLLM still preserves AR scaling at a fraction of the cost. On coding tasks, it trades slightly worse pass@$k$ scaling in favor of better pass@NFE scaling. Finally, it again consistently matches or outperforms Fast-dLLM and TCT. We additionally report the exact values from LLaDA-8B-Instruct and Dream-v0-Instruct-7B in Table \ref{tab:passk-llada} and \ref{tab:passk-dream}.

\begin{table}[ht]
\centering
\small
\setlength{\tabcolsep}{4pt}
\resizebox{\linewidth}{!}{%
\begin{tabular}{lcccccccccccc}
\toprule
\multirow{2}{*}{Method} & \multicolumn{3}{c}{GSM8K} & \multicolumn{3}{c}{MATH-500} & \multicolumn{3}{c}{HumanEval} & \multicolumn{3}{c}{MBPP} \\
\cmidrule(lr){2-4} \cmidrule(lr){5-7} \cmidrule(lr){8-10} \cmidrule(lr){11-13}
 & Pass@1 & Pass@64 & NFE & Pass@1 & Pass@64 & NFE & Pass@1 & Pass@64 & NFE & Pass@1 & Pass@64 & NFE \\
\midrule
Fork-dLLM & \textbf{80.9} & 98.3 & 67.8 & 31.7 & \textbf{76.6} & 92.3 & 31.7 & 78.6 & 108.4 & 33.8 & 74.1 & 75.8 \\
Fast-dLLM & 80.1 & 97.3 & 58.3 & \textbf{32.1} & 73.4 & 76.0 & \textbf{35.7} & 75.0 & 87.9 & \textbf{36.7} & 73.3 & 60.5 \\
TCT & 75.1 & 98.3 & \textbf{51.5} & 28.5 & 73.4 & \textbf{67.3} & 31.4 & 73.2 & \textbf{74.1} & 33.7 & 72.1 & \textbf{54.0} \\
AR & \textbf{80.9} & \textbf{99.0} & 227.9 & 31.4 & 74.4 & 243.3 & 34.0 & \textbf{78.8} & 242.3 & 34.8 & \textbf{74.7} & 192.3 \\
\bottomrule
\end{tabular}%
}
\vspace{10pt}
\caption{Pass@$1$ and pass@$64$ results on LLaDA-8B-Instruct. Best per column in bold.}
\label{tab:passk-llada}
\end{table}

\begin{table}[ht]
\centering
\small
\setlength{\tabcolsep}{4pt}
\resizebox{\linewidth}{!}{%
\begin{tabular}{lcccccccccccc}
\toprule
\multirow{2}{*}{Method} & \multicolumn{3}{c}{GSM8K} & \multicolumn{3}{c}{MATH-500} & \multicolumn{3}{c}{HumanEval} & \multicolumn{3}{c}{MBPP} \\
\cmidrule(lr){2-4} \cmidrule(lr){5-7} \cmidrule(lr){8-10} \cmidrule(lr){11-13}
 & Pass@1 & Pass@64 & NFE & Pass@1 & Pass@64 & NFE & Pass@1 & Pass@64 & NFE & Pass@1 & Pass@64 & NFE \\
\midrule
Fork-dLLM & 81.7 & \textbf{99.3} & 78.0 & 38.0 & \textbf{83.4} & 97.3 & 46.8 & \textbf{92.8} & 63.9 & 50.6 & 85.6 & 43.3 \\
Fast-dLLM & 81.2 & \textbf{99.3} & 67.4 & \textbf{38.4} & 82.8 & 82.6 & 45.9 & 90.2 & 51.2 & 51.3 & 83.4 & \textbf{34.7} \\
TCT & 74.2 & \textbf{99.3} & \textbf{59.9} & 33.7 & 80.0 & \textbf{73.1} & 39.9 & 89.6 & \textbf{47.6} & 45.6 & 84.0 & 35.6 \\
AR & \textbf{82.4} & 99.0 & 198.5 & 37.9 & 83.0 & 223.1 & \textbf{51.1} & 92.7 & 131.2 & \textbf{53.0} & \textbf{88.4} & 71.9 \\
\bottomrule
\end{tabular}%
}
\vspace{10pt}
\caption{Pass@$1$ and pass@$64$ results on Dream-v0-Instruct-7B. Best per column in bold.}
\label{tab:passk-dream}
\end{table}

\begin{figure}[t]
    \centering
    \includegraphics[width=0.5\textwidth]{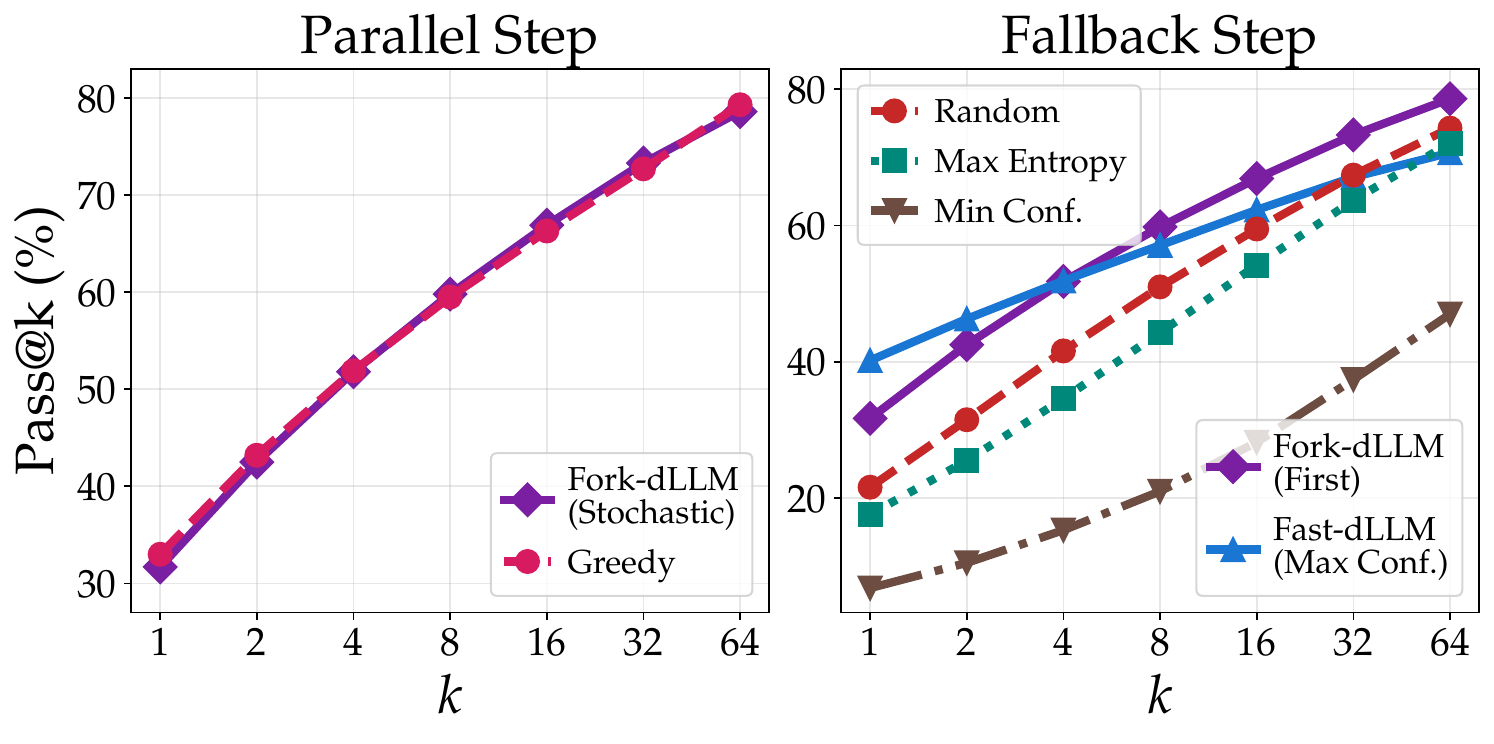}
    \caption{Pass@$k$ curves of Fork-dLLM and its ablations on HumanEval using LLaDA-8B-Instruct.}
    \label{fig:fork-ablations-k}
\end{figure}

\begin{figure}[h]
    \centering
    \includegraphics[width=0.6\textwidth]{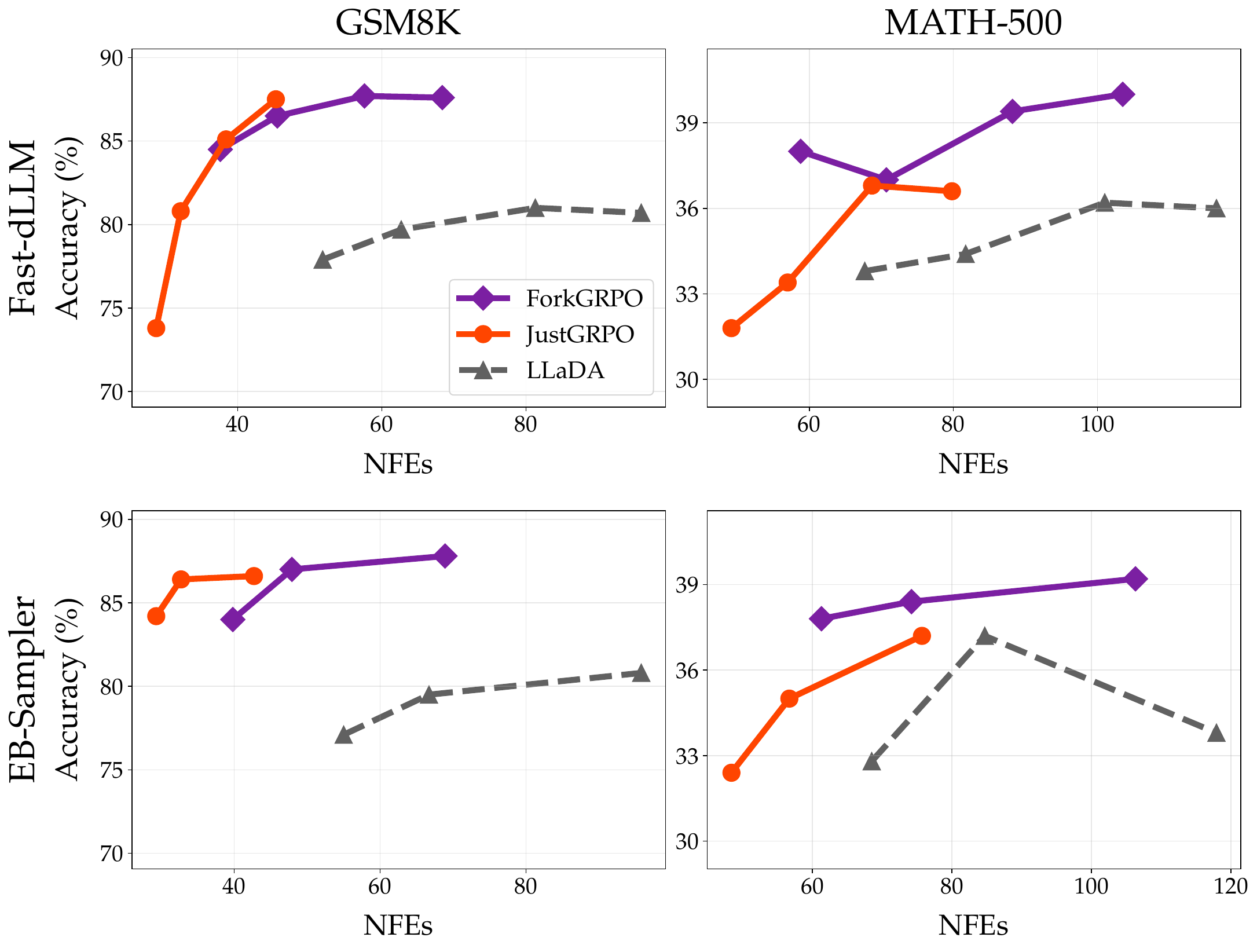}
    \caption{Accuracy versus NFEs of the final ForkGRPO and JustGRPO checkpoints on GSM8K (left) and MATH-500 (right), evaluated with Fast-dLLM (top) and EB (bottom) across thresholds. The JustGRPO checkpoint is taken from \cite{ni2026flexibility}.}
    \label{fig:checkpoints-all}
\end{figure}

\end{document}